\documentclass[letterpaper, 10 pt, conference]{IEEEtran}  

\IEEEoverridecommandlockouts                               
\usepackage{cite}
\usepackage{amsmath,amssymb,amsfonts}
\usepackage{algorithmic}
\usepackage{graphicx}
\usepackage{textcomp}
\usepackage{xcolor}
\usepackage{enumerate}
\usepackage[linesnumbered,ruled,lined]{algorithm2e}
\usepackage{booktabs}
\usepackage{threeparttable}
\usepackage{multirow}
\usepackage{stfloats}
\usepackage{subfigure}
\usepackage{graphicx}
\usepackage{rotating}
\usepackage[colorlinks,linkcolor=blue]{hyperref}
\usepackage{bm}  

\usepackage{multirow}
\usepackage[normalem]{ulem}
\useunder{\uline}{\ul}{}
\usepackage{tabularx}  

\usepackage{geometry}
\title{\LARGE \bf
RGBDS-SLAM: A RGB-D Semantic Dense SLAM Based on 3D Multi Level Pyramid Gaussian Splatting}

\author{Zhenzhong Cao$^{1}$, Chenyang Zhao$^{1}$, Qianyi Zhang$^{1}$, Jinzheng Guang$^{1}$, Yinuo Song$^{1}$, Jingtai Liu$^{1*}$
	\thanks{$^*$The corresponding author of this paper. }
        \thanks{This work is supported by the National Natural Science Foundation of China under Grant 62173189.}
	\thanks{$^{1}$The authors are with the Institute of Robotics and Automatic Information System, Nankai University, Tianjin 300350, China; Tianjin Key Laboratory of Intelligent Robotics, Tianjin 300350, China; and also with TBI center, Nankai University, Tianjin 300350, China (e-mail: caozhenzhong@mail.nankai.edu.cn; liujt@nankai.edu.cn).}
}

\begin{document}

\maketitle
\thispagestyle{empty}
\pagestyle{empty}

\begin{abstract}
 High-fidelity reconstruction is crucial for dense SLAM. Recent popular methods utilize 3D gaussian splatting (3D GS) techniques for RGB, depth, and semantic reconstruction of scenes. However, these methods ignore issues of detail and consistency in different parts of the scene. To address this, we propose RGBDS-SLAM, a RGB-D semantic dense SLAM system based on 3D multi-level pyramid gaussian splatting, which enables high-fidelity dense reconstruction of scene RGB, depth, and semantics. In this system, we introduce a 3D multi-level pyramid gaussian splatting method that restores scene details by extracting multi-level image pyramids for gaussian splatting training, ensuring consistency in RGB, depth, and semantic reconstructions. Additionally, we design a tightly-coupled multi-features reconstruction optimization mechanism, allowing the reconstruction accuracy of RGB, depth, and semantic features to mutually enhance each other during the rendering optimization process. Extensive quantitative, qualitative, and ablation experiments on the Replica and ScanNet public datasets demonstrate that our proposed method outperforms current state-of-the-art methods, which achieves great improvement by
11.13\% in PSNR and 68.57\% in LPIPS. The open-source code will be available at: \href{https://github.com/zhenzhongcao/RGBDS-SLAM}{https://github.com/zhenzhongcao/RGBDS-SLAM}.
\end{abstract}

\section{INTRODUCTION}
Visual SLAM is a fundamental problem in the field of robotics, aimed at solving the problem of simultaneously locating a robot and constructing a map of its surrounding environment. Dense mapping is an important component of visual SLAM; on the one hand, it enables the robot to perceive its surroundings more comprehensively, and on the other hand, it provides a foundational map for downstream tasks such as grasping, manipulation, and interaction. However, traditional dense visual SLAM \cite{whelan2016elasticfusion, engel2017direct, mur2017orb, scona2018staticfusion, zhang2020flowfusion, campos2021orb} relies solely on point clouds to reconstruct scenes, and due to the limited number of points and their discontinuous distribution, it faces significant bottlenecks and cannot achieve high-fidelity reconstructions of the environment.

With the advent of NeRF (Neural Radiance Fields)\cite{mildenhall2021nerf}, scene representation based on implicit neural radiance fields has gradually become popular. Through training, the reconstruction accuracy has significantly improved, and many approaches have incorporated NeRF into SLAM\cite{zhu2022nice, yang2022vox, wang2023co, johari2023eslam, sandstrom2023point, haghighi2023neural, li2023dns, zhu2024sni}, achieving high-precision RGB, depth, and semantic Reconstructions. However, NeRF itself suffers from issues such as long training times and slow rendering speeds, meaning that NeRF-based SLAM solutions cannot run in real time, which contradicts the original goal of SLAM.

3D GS\cite{kerbl20233d} technology, with its efficient optimization framework and real-time rendering capability, improves upon the shortcomings of NeRF. As a result, many 3D GS-based SLAM \cite{keetha2024splatam, yan2024gs, matsuki2024gaussian, yugay2023gaussian, huang2024photo, ji2024neds, zhu2024semgauss, li2025sgs} solutions have emerged. However, these methods typically train using only raw image features, which are insufficient to fully capture the fine-grained details of certain scene parts, leading to poor reconstruction consistency. Moreover, when performing multi-feature reconstruction, these approaches do not effectively fuse and optimize the features through reasonable constraints, preventing them from mutually enhancing each other.

To address the key issues of insufficient detail restoration, poor reconstruction consistency, ineffective fusion of multi-feature information, and real-time challenges in reconstruction, we propose the RGBDS-SLAM algorithm in this paper. First, we introduce a 3D multi-level pyramid gaussian splatting method, which constructs a multi-level image pyramid to extract rich detail information at different resolution levels and perform gaussian splatting training. This method significantly improves the scene's detail restoration capability, and through stepwise optimization across levels, it ensures effective global consistency during reconstruction, providing a solid foundation for precise restoration of complex scenes. Second, we design a tightly coupled multi-features reconstruction optimization mechanism, which reasonably couples RGB, depth, and semantic features through various constraints. In the rendering optimization process, these three features collaborate and promote each other. Semantic information enhances depth understanding, depth information supports semantic refinement, and at the same time, the realism and consistency of RGB rendering are optimized, thereby comprehensively improving the accuracy and reliability of reconstruction. Finally, we develop a complete RGB-D Semantic Dense SLAM system, achieving high-quality dense reconstruction of scene RGB color, depth information, and semantic color. This system is based on the current classic ORB-SLAM3 algorithm\cite{campos2021orb}, capable of processing complex scenes in real time and meeting the dual requirements of speed and accuracy for online applications.

\begin{figure*}[ht]
	\centering
	\includegraphics[scale=0.51]{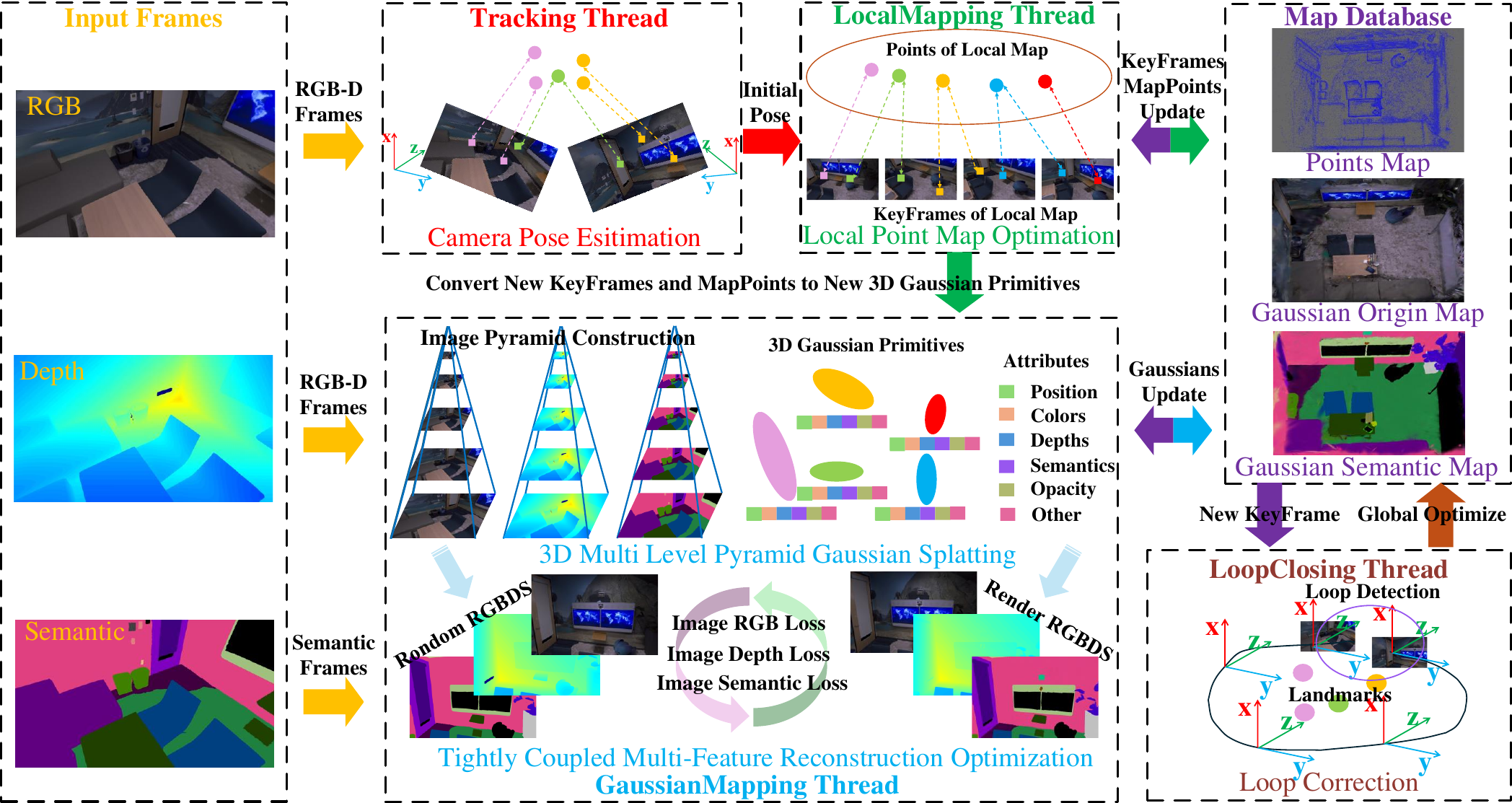}
	\caption{Overview of the proposed RGBDS-SLAM. Our method is an enhancement of ORB-SLAM3\cite{campos2021orb}, taking RGB, depth, and semantic frames as input and outputting a map database with the point map, gaussian origin map, and gaussian semantic map. It consists of four threads: Tracking, LocalMapping, GaussianMapping, and LoopClosing.}
	\label{systemframework}
	\vspace{-5mm}
\end{figure*}

\textbf{The main contributions of this work are as follows:}
\begin{itemize}
	\item We introduce a \textbf{3D Multi-Level Pyramid Gaussian Splatting (MLP-GS)} method, which extracts multi-level image pyramids for gaussian splatting training, restoring scene details and ensuring consistency during reconstruction.
	\item We design a \textbf{Tightly Coupled Multi-Features Reconstruction Optimization(TCMF-RO)} mechanism, which promotes mutual improvement of RGB, depth, and semantic map reconstruction accuracy during the optimization rendering process.
	\item We develop a \textbf{complete RGB-D Semantic Dense SLAM system} capable of high-quality dense reconstruction of scene RGB, depth, and semantic information, and the system can operate in real time. We will also open source our code once the paper is accepted.
\end{itemize}

\section{Related Work}

\subsection{NeRF-based SLAM}
The development in neural implicit representations, particularly those based on NeRF, have significantly enhanced the performance of SLAM systems. Among them, NICE-SLAM\cite{zhu2022nice} is the first solution to combine NeRF and SLAM, which incorporates multi-level local information by introducing a hierarchical scene representation, enabling efficient map construction and robust tracking. However, NICE-SLAM suffers from computational efficiency issues. Therefore, \cite{yang2022vox}, \cite{wang2023co}, \cite{johari2023eslam} and \cite{sandstrom2023point} have introduced voxel-based neural representations, coordinate and sparse parameters, hybrid representation of signed distance fields (SDF) and neural point cloud respectively to optimize and improve computational efficiency. The above solutions do not consider semantic mapping, so based on these solutions, NIDS-SLAM\cite{haghighi2023neural} introduce a novel approach for dense 3D semantic segmentation, based on 2D semantic color information of keyframes, which are able to accurately learn the dense 3D semantics of the scene online while simultaneously learning geometry. However, this work does not integrate semantic with other features of the environment, such as geometry and appearance. Therefore, DNS-SLAM\cite{li2023dns} integrates multi-view geometry constraints with image-based feature extraction to improve appearance details and to output color, density, and semantic class information. SNI-SLAM\cite{zhu2024sni} introduce cross-attention based feature fusion to incorporate semantic, appearance, and geometry features, thus improving the accuracy of mapping, tracking, and semantic segmentaion. Although these NeRF-based SLAM schemes achieve high-quality reconstruction effects, they suffer from poor scalability, low efficiency and poor real-time performance due to NeRF.

\subsection{3D GS-based SLAM}
The emergence of 3D GS have led to significant advancements in both general and semantic SLAM systems. \cite{keetha2024splatam, yan2024gs, matsuki2024gaussian, yugay2023gaussian} pioneered the introduction of 3D GS technology into SLAM systems, which are all committed to continuously expanding and optimizing gaussian map parameters in the incremental process of SLAM to achieve high-fidelity incremental reconstruction of scenes. However, their camera tracking modules all rely on gradient optimization of image loss, so the real-time performance of the systems is relatively poor. Photo-SLAM\cite{huang2024photo} introduces ORB-SLAM3 as the basic framework to improve this problem. None of the above solutions performs semantic mapping of the scene. Therefore, based on these solutions, SGS-SLAM\cite{li2025sgs} proposes to employ multi-channel optimization during the mapping process, integrating appearance, geometric, and semantic constraints with keyframe optimization to enhance reconstruction quality. NEDS-SLAM\cite{ji2024neds} propose a spatially consistent feature fusion model to reduce the effect of erroneous estimates from pre-trained segmentation head on semantic reconstruction, achieving robust 3D semantic gaussian mapping. Although these 3D GS-based SLAM schemes achieve high-efficiency and high-precision dense reconstruction, they do not restore enough scene details, have poor consistency, and have low coupling of multi-feature information.

\section{RGBDS-SLAM Algorithmn}
\subsection{Overall System Framework}
 The Fig.\ref{systemframework} illustrates the overall framework of the proposed RGBDS-SLAM, which is based on ORB-SLAM3\cite{campos2021orb}. The system takes RGB, depth, and semantic frames as input data and outputs a map database containing the point map, gaussian origin map, and gaussian semantic map. It primarily consists of four threads: \textit{Tracking Thread}, \textit{LocalMapping Thread}, \textit{GaussianMapping Thread}, and \textit{LoopClosing Thread}. The specific data flow between these threads is as follows:

\textit{Tracking Thread}: Receives RGB-D frame data and estimates the camera pose for the current frame.

\textit{LocalMapping Thread}: Receives the initial pose provided by the \textit{Tracking Thread}, determines whether a new keyframe can be created, and if so, creates new keyframes and map points, optimizes the local map, and updates the point cloud map.

\textit{GaussianMapping Thread}: Receives the new keyframe and map point data created by the \textit{LocalMapping Thread}, converts it into 3D gaussian primitives (including position, color, semantics, depth, opacity, etc.), then performs the 3D multi-level pyramid gaussian splatting operation. Finally, the gaussian origin map and gaussian semantic map are updated through the tightly coupled multi-features reconstruction optimization mechanism.

\textit{Loop Closing Thread}: Accepts new keyframe data from the map, performs loop closure, and if a loop is detected, executes global optimization and updates the entire map.

\subsection{3D Gaussian Primitives Representation}
We define that each 3D gaussian primitive includes position, shape, RGB color, depth value, and semantic color information. Referring to the operation in \cite{straub2019replica} that simplifies the gaussian parameters by reducing the shape component (transforming the covariance matrix from anisotropic to isotropic), we can define the expression for the influence of a 3D gaussian primitive on other spatial locations as follows:
\begin{equation}
{g^{3D}}(\bm{x}) = o\exp \left( - \frac{{{{\left\| \bm{x} - \bm{\mu} \right\|}^2}}}{{2{r^2}}}\right)
\end{equation}where $\bm{\mu}$ is the position of the 3D gaussian primitive, $\bm{r}$ is the shape, $\bm{x}$ is the spatial location, and $o$ is the opacity. 

As for data preparation of gaussian splatting, we convert the parameters in (1) into 2D using the camera's intrinsic parameters $K \in {R^{3 \times 3}}$ (symmetric matrix), focal length $f$, and extrinsic parameters ${T_{cw}} \in {R^{3 \times 4}}$ (the transformation from world coordinates to camera coordinates):
\begin{equation}
{\bm{\mu} ^{2D}} = K\frac{{{T_{cw}}\bm{\mu} }}{d},{r^{2D}} = \frac{{fr}}{d},d = {({T_{c,w}}\bm{\mu})_z}
\end{equation}

By using the above equation, we project the 3D gaussian primitive onto the image plane to obtain a 2D gaussian primitive. We can then define the expression for the influence of the 2D gaussian primitive on other image pixels as follows:
\begin{equation}
{g^{2D}}(\bm{p}) = o\exp ( - \frac{{{{\left\| {\bm{p} - {\bm{\mu} ^{2D}}} \right\|}^2}}}{{2{{({r^{2D}})}^2}}})
\end{equation}

Using the above equation, we can proceed with the subsequent gaussian splatting operations. Additionally, for each 3D gaussian primitive, we convert its RGB color and semantic color information into multi-dimensional feature vectors $\bm{r}$ and $\bm{s}$ using the SH (Spherical Harmonics) method to represent them.

\subsection{3D Multi-Level Pyramid Gaussian Splatting}
Unlike the standard 3D gaussian splatting process, we refer to the progressive training process proposed in \cite{liu2020neural, sun2022direct, takikawa2021neural, li2023neuralangelo, xiangli2022bungeenerf} and introduce a 3D multi-level pyramid gaussian splatting. In this process, the resolution of various feature images (RGB, depth, and semantic images) is gradually increased during training. This not only reduces training time and difficulty, but also allows for the gradual reconstruction of multi-scale information for different features at different resolutions. 

\begin{figure}[ht]
	\centering
	\includegraphics[scale=0.58]{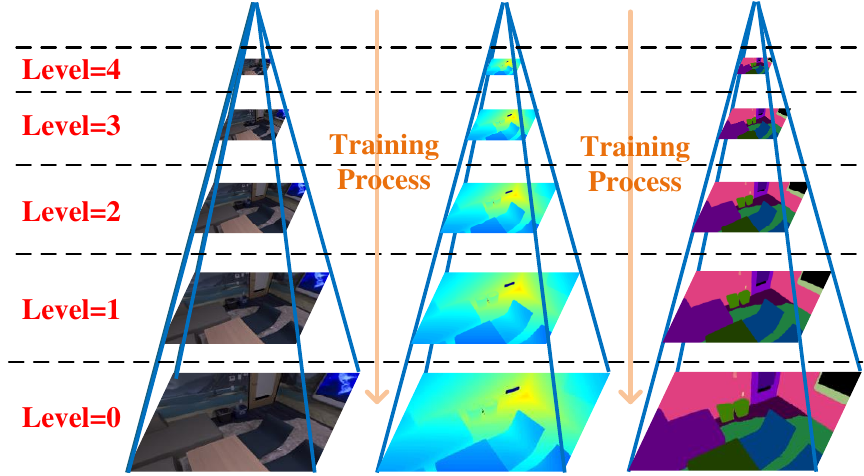}
	\caption{Multi level image pyramid construction. During the training process, it is carried out from top to bottom, with the resolution of the image gradually increasing. First, low resolution is used for quick initialization, and then the details are gradually improved.}
	\label{semantic}
\end{figure}

Therefore, we construct an n-layer image pyramid for RGB, depth, and semantic images.

The \(i\)-th layer of the RGB pyramid image can be represented as:
\begin{equation}
I_r^{gt}(i) = PyramidImageExtrcation(I_{RGB}^{gt},i)
\end{equation}

The \(i\)-th layer of the depth pyramid image can be represented as:
\begin{equation}
I_d^{gt}(i) = PyramidImageExtrcation(I_{depth}^{gt},i)
\end{equation}

The \(i\)-th layer of the semantic pyramid image can be represented as:
\begin{equation}
I_s^{gt}(i) = PyramidImageExtrcation(I_{semantic}^{gt},i)
\end{equation}

During the training process, to ensure comprehensive training for each viewpoint and each layer of the image pyramid, in each iteration, we randomly select a set of multi-feature images $\{ I_r^{gt}(i),I_d^{gt}(i),I_s^{gt}(i)\}$. We extract all relevant information for that viewpoint (such as pose, image size, etc.), and based on this information, we perform rendering operations for RGB, depth, and semantic images, referring to the rendering formula proposed in \cite{kerbl20233d}.

We perform RGB rendering operation using:
\begin{equation}
R(\bm{p}) = \sum\limits_{i \in N} {{r_i}} g_i^{2D}(\bm{p})\prod\limits_{j = 1}^{i - 1} {(1 - g_j^{2D}(\bm{p}))}
\end{equation}

We perform depth rendering operation using:
\begin{equation}
D(\bm{p}) = \sum\limits_{i \in N} {{d_i}} g_i^{2D}(\bm{p})\prod\limits_{j = 1}^{i - 1} {(1 - g_j^{2D}(\bm{p}))} 
\end{equation}

We perform semantic rendering operation using:
\begin{equation}
S(\bm{p}) = \sum\limits_{i \in N} {{s_i}} g_i^{2D}(\bm{p})\prod\limits_{j = 1}^{i - 1} {(1 - g_j^{2D}(\bm{p}))}
\end{equation}
where, the set N represents the sorted 2D gaussian primitives required to render the RGB, depth and semantic of the $\bm{p}$ pixel, and the cumulative multiplication operation represents the cumulative effect of the previous 2D gaussian primitives on the current one.

Through our proposed MLP-GS progressive training process, we can gradually restore the scene details to the maximum extent.

\subsection{Tightly Coupled Multi-Feature Reconstruction Optimization}
In the previous section, we performed MLP-GS operations on the 3D gaussian primitives in the map, resulting in a set of rendered images $\{ I_r^{rd}(i),I_d^{rd}(i),I_s^{rd}(i)\} $. This is the forward rendering process of gaussian splatting. We now need to compute the loss between the rendered images and the ground truth images and perform backpropagation to optimize the 3D gaussian primitives in the map.

Referring to the calculation of L1 loss and SSIM loss for rendered images and the groundtruth images in \cite{li2025sgs}, we perform a similar loss calculation on the rendered images $\{ I_r^{gt}(i),I_d^{gt}(i),I_s^{gt}(i)\}$ of the i-th pyramid perspective obtained in the previous section.

For RGB images, we consider L1 and SSIM loss:
\begin{equation}
{L_r}(i) = (1 - {\lambda _r})\left| {I_r^{rd}(i) - I_r^{gt}(i)} \right| + {\lambda _r}SSIM(I_r^{rd}(i),I_r^{gt}(i))
\end{equation}

For depth images, we only consider L1 loss:
\begin{equation}
{L_d}(i) = \left| {I_d^{rd}(i) - I_d^{gt}(i)} \right|
\end{equation}

For semantic images, we similarly consider L1 and SSIM loss:
\begin{equation}
{L_s}(i) = (1 - {\lambda _s})\left| {I_s^{rd}(i) - I_s^{gt}(i)} \right| + {\lambda _s}SSIM(I_s^{rd}(i),I_s^{gt}(i))
\end{equation}

Finally, we tightly couple multiple features into a reconstruction optimization framework to perform joint optimization:
\begin{equation}
{L_{reconstruction}}(i) = {L_r}(i) + {L_d}(i) + {L_s}(i)
\end{equation}

Through the proposed TCMF-RO, which couples multiple features within a single framework, the RGB, depth, and semantic features in the 3D gaussian primitives can promote and enhance each other during optimization.

\begin{table*}[]
\begin{center}
\caption{Quantitative comparison of RGB reconstruction quality between our method and baselines on 8 sequences of Replica dataset.}\label{t1}
\resizebox{\linewidth}{!}{
\begin{tabular}{cccccccccccc}
\hline
\multicolumn{2}{c}{Method}                                       & Metric & office0        & office1        & office2        & office3        & office4        & room0          & room1          & room2          & avg            \\ \hline
\multirow{12}{*}{NeRF-based SLAM}  & \multirow{3}{*}{NICE-SLAM\cite{zhu2022nice}}  & PSNR↑  & 29.07          & 30.34          & 19.66          & 22.23          & 24.94          & 22.12          & 22.47          & 24.52          & 24.42          \\
                                   &                             & SSIM↑  & 0.874          & 0.886          & 0.797          & 0.801          & 0.856          & 0.689          & 0.757          & 0.814          & 0.809          \\
                                   &                             & LPIPS↓ & 0.229          & 0.181          & 0.235          & 0.209          & 0.198          & 0.330          & 0.271          & 0.208          & 0.233          \\ \cline{2-12} 
                                   & \multirow{3}{*}{Vox-Fusion\cite{yang2022vox}} & PSNR↑  & 27.79          & 29.83          & 20.33          & 23.47          & 25.21          & 22.39          & 22.36          & 23.92          & 24.41          \\
                                   &                             & SSIM↑  & 0.857          & 0.876          & 0.794          & 0.803          & 0.847          & 0.683          & 0.751          & 0.798          & 0.801          \\
                                   &                             & LPIPS↓ & 0.241          & 0.184          & 0.243          & 0.213          & 0.199          & 0.303          & 0.269          & 0.234          & 0.236          \\ \cline{2-12} 
                                   & \multirow{3}{*}{Co-SLAM\cite{wang2023co}}    & PSNR↑  & 34.14          & 34.87          & 28.43          & 28.76          & 30.91          & 27.27          & 28.45          & 29.06          & 30.24          \\
                                   &                             & SSIM↑  & 0.961          & 0.969          & 0.938          & 0.941          & 0.955          & 0.910          & 0.909          & 0.932          & 0.939          \\
                                   &                             & LPIPS↓ & 0.209          & 0.196          & 0.258          & 0.229          & 0.236          & 0.324          & 0.294          & 0.266          & 0.252          \\ \cline{2-12} 
                                   & \multirow{3}{*}{ESLAM\cite{wang2023co}}      & PSNR↑  & 33.71          & 30.20          & 28.09          & 28.77          & 29.71          & 25.32          & 27.77          & 29.08          & 29.08          \\
                                   &                             & SSIM↑  & 0.960          & 0.923          & 0.943          & 0.948          & 0.945          & 0.875          & 0.902          & 0.932          & 0.929          \\
                                   &                             & LPIPS↓ & 0.184          & 0.228          & 0.241          & 0.196          & 0.204          & 0.313          & 0.298          & 0.248          & 0.239          \\ \hline
\multirow{15}{*}{3D GS-based SLAM} & \multirow{3}{*}{SplaTAM\cite{keetha2024splatam}}    & PSNR↑  & 38.26          & 39.17          & 31.97          & 29.70          & 31.81          & 32.86          & 33.89          & 35.25          & 34.11          \\
                                   &                             & SSIM↑  & 0.98           & 0.98           & 0.97           & 0.95           & 0.95           & 0.98           & 0.97           & 0.98           & 0.970          \\
                                   &                             & LPIPS↓ & 0.09           & 0.09           & 0.10           & 0.12           & 0.15           & 0.07           & 0.10           & 0.08           & 0.100          \\ \cline{2-12} 
                                   & \multirow{3}{*}{Photo-SLAM\cite{huang2024photo}} & PSNR↑  & 38.48          & 39.09          & {\ul 33.03}          & {\ul 33.79}          & {\ul 36.02}          & 30.72          & 33.51          & 35.03          & {\ul 34.96}    \\
                                   &                             & SSIM↑  & 0.964          & 0.961          & 0.938          & 0.938          & 0.952          & 0.899          & 0.934          & 0.951          & 0.942          \\
                                   &                             & LPIPS↓ & {\ul 0.050}          & {\ul 0.047}          & {\ul 0.077}          & {\ul 0.066}          & {\ul 0.054}          & 0.075          & {\ul 0.057}          & {\ul 0.043}          & {\ul 0.059}    \\ \cline{2-12} 
                                   & \multirow{3}{*}{NEDS-SLAM\cite{ji2024neds}}  & PSNR↑  & /              & /              & /              & /              & /              & /              & /              & /              & 34.76          \\
                                   &                             & SSIM↑  & /              & /              & /              & /              & /              & /              & /              & /              & 0.962          \\
                                   &                             & LPIPS↓ & /              & /              & /              & /              & /              & /              & /              & /              & 0.088          \\ \cline{2-12} 
                                   & \multirow{3}{*}{SGS-SLAM\cite{li2025sgs}}   & PSNR↑  & {\ul 38.54}          & {\ul 39.20}          & 32.90          & 32.05          & 32.75          & {\ul 32.50}          & {\ul 34.25}          & {\ul 35.10}          & 34.66          \\
                                   &                             & SSIM↑  & \textbf{0.984} & \textbf{0.982} & \textbf{0.965} & \textbf{0.966} & {\ul 0.949}    & \textbf{0.976} & \textbf{0.978} & \textbf{0.982} & \textbf{0.973} \\
                                   &                             & LPIPS↓ & 0.086          & 0.087          & 0.101          & 0.115          & 0.148          & {\ul 0.070}          & 0.094          & 0.070          & 0.096          \\ \cline{2-12} 
                                   & \multirow{3}{*}{RGBDS-SLAM(Ours)} & PSNR↑  & \textbf{42.46} & \textbf{42.57} & \textbf{35.80} & \textbf{36.53} & \textbf{39.47} & \textbf{35.77} & \textbf{38.59} & \textbf{39.58} & \textbf{38.85} \\
                                   &                             & SSIM↑  & {\ul 0.981}    & {\ul 0.976}    & {\ul 0.959}    & {\ul 0.958}    & \textbf{0.969} & {\ul 0.955}    & {\ul 0.968}    & {\ul 0.973}    & {\ul 0.967}    \\
                                   &                             & LPIPS↓ & \textbf{0.023} & \textbf{0.029} & \textbf{0.052} & \textbf{0.046} & \textbf{0.034} & \textbf{0.037} & \textbf{0.029} & \textbf{0.027} & \textbf{0.035} \\ \hline
\end{tabular}%
}
\footnotesize{ \\ / indicates that the paper does not provide relevant data, \textbf{bold data} indicates optimal data, and {\ul underlined data} indicates suboptimal data.}
\end{center}
\vspace{-3mm}
\end{table*}

\begin{table*}[]
\centering
\caption{Quantitative comparison of average results on Depth, ATE, and FPS metrics between our method and baselines on 8 sequences of Replica dataset.}\label{t2}
\resizebox{\linewidth}{!}{
\begin{tabular}{ccccccc}
\hline
\multicolumn{2}{c}{Method}                     & Depth(cm)↓     & ATE Mean (cm)↓ & ATE RMSE (cm)↓ & Tracking FPS↑  & Mapping FPS↑   \\ \hline
\multirow{5}{*}{NeRF-based SLAM}  & NICE-SLAM\cite{zhu2022nice}  & 1.903          & 1.795          & 2.503          & 13.70          & 0.20           \\
                                  & Vox-Fusion\cite{yang2022vox} & 2.913          & 1.027          & 1.473          & 2.11           & 2.17           \\
                                  & Co-SLAM\cite{wang2023co}    & 1.513          & 0.935          & 1.059          & 17.24          & {\ul 10.20}          \\
                                  & ESLAM\cite{johari2023eslam}      & 0.945          & 0.545          & 0.678          & 18.11          & 3.62           \\
                                  & SNI-SLAM\cite{zhu2024sni}   & 0.766          & {\ul 0.397}    & 0.456          & 16.03          & 2.48           \\ \hline
\multirow{5}{*}{3D GS-based SLAM} & SplaTAM\cite{keetha2024splatam}    & 0.490          & /              & 0.360          & 5.26           & 3.03           \\
                                  & Photo-SLAM\cite{huang2024photo} & /              & /              & 0.604          & \textbf{42.49}          & /              \\
                                  & NEDS-SLAM\cite{ji2024neds}  & 0.470          & /              & {\ul 0.354}    & /              & /              \\
                                  & SGS-SLAM\cite{li2025sgs}   & {\ul 0.356}          & \textbf{0.327} & \textbf{0.412} & 5.27           & 3.52           \\
                                  & RGBDS-SLAM(Ours) & \textbf{0.342} & 0.499          & 0.589          & {\ul 29.55} & \textbf{32.22} \\ \hline
\end{tabular}%
}
\end{table*}

\begin{figure*}[ht]
	\centering
	\includegraphics[scale=0.47]{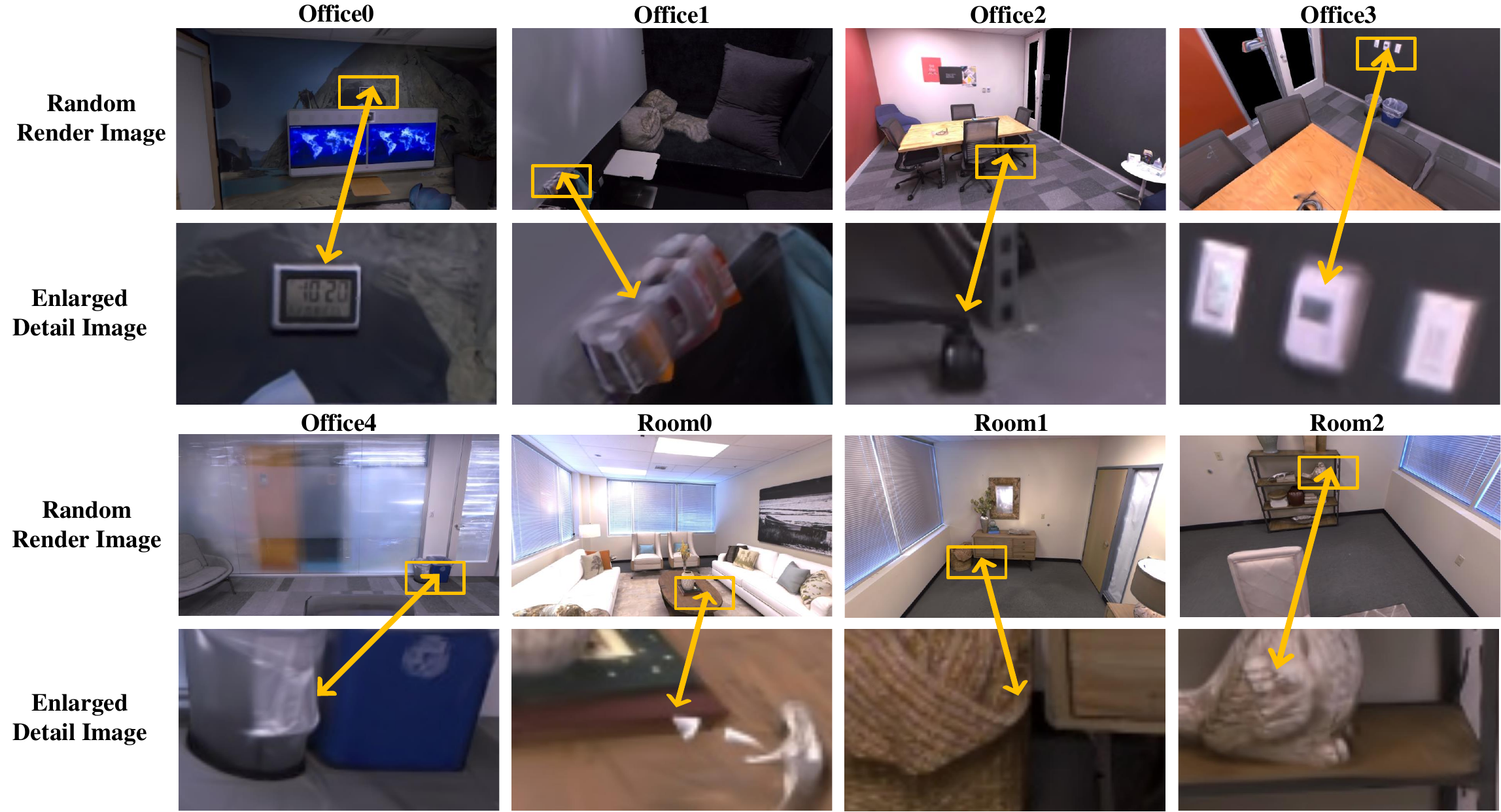}
	\caption{Qualitative performance of our proposed method on RGB image rendering details from 8 sequences of the Replica dataset is shown. The first and third rows display the randomly rendered RGB images from the 8 sequences, while the second and fourth rows show the corresponding zoomed-in details. The regions of interest in the zoomed-in images are indicated with orange boxes and arrow lines to highlight the magnified details.}
	\label{RGB}
\end{figure*}

\begin{figure*}[ht]
	\centering
	\includegraphics[scale=0.47]{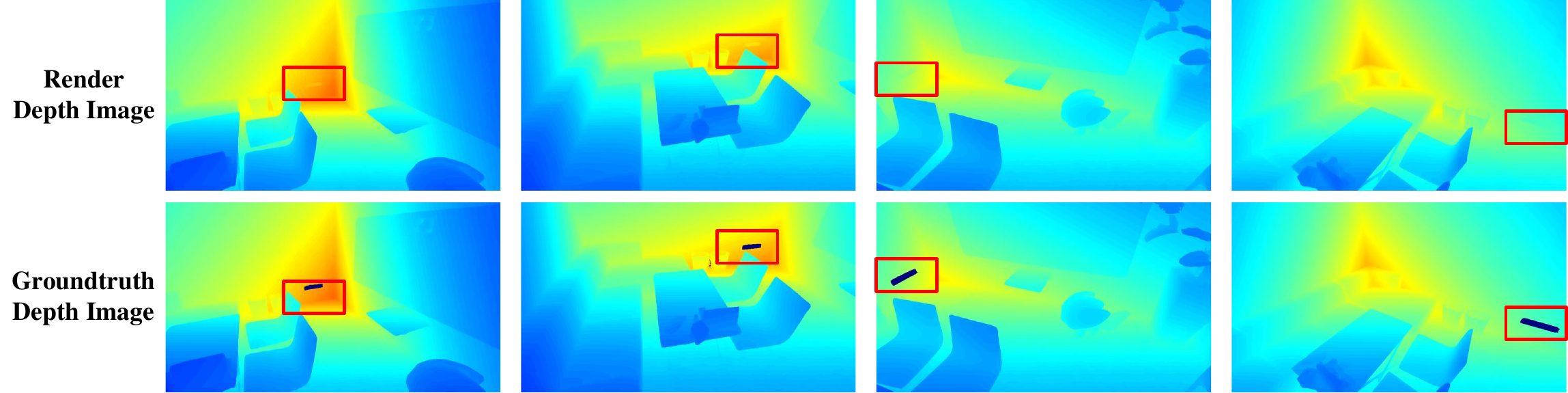}
	\caption{Qualitative comparison of rendered depth images and groundtruth depth images of our method on office0 sequence of Replica dataset. The first row is the randomly rendered depth images, and the second row is the corresponding groundtruth depth images. The red boxes indicate the differences. The red boxes on the groundtruth depth indicate the areas with missing depth.}
	\label{depth}
\end{figure*}

\begin{figure*}[ht]
	\centering
	\includegraphics[scale=0.46]{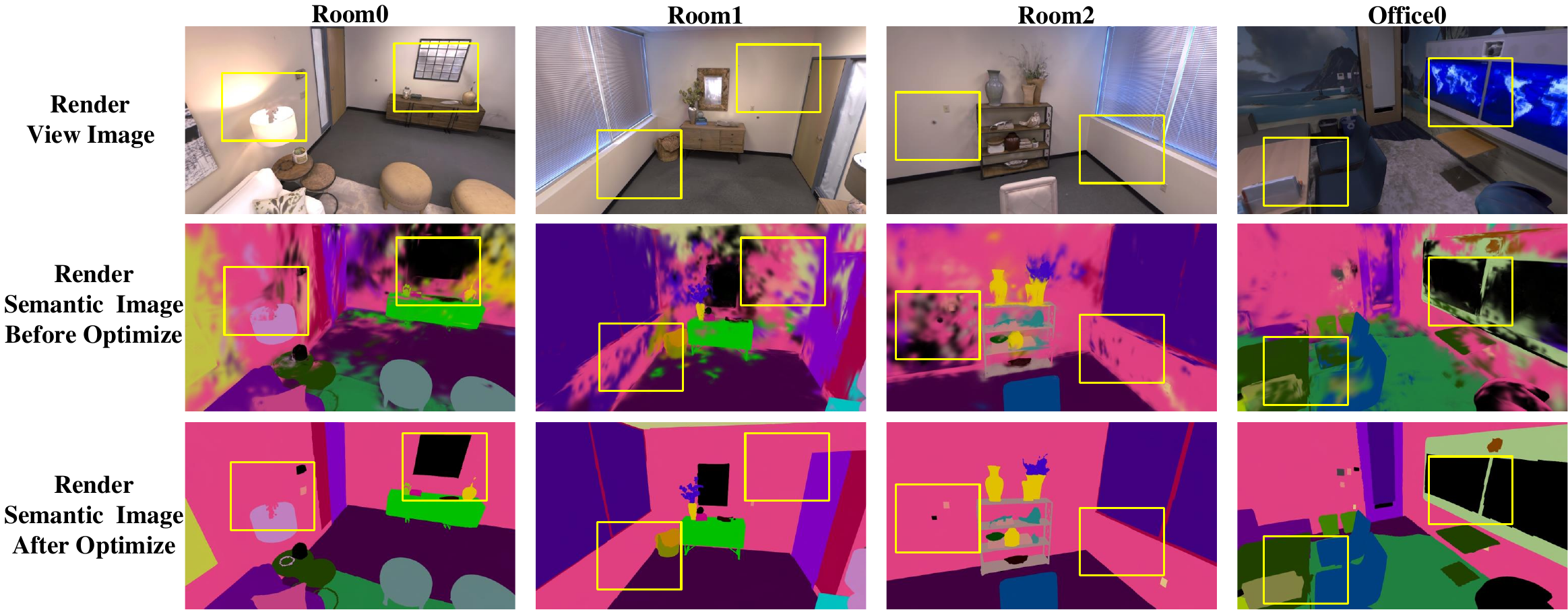}
	\caption{Qualitative comparison of semantic image rendering of our method on four sequences of Replica dataset. The first row is the RGB image rendered from a random perspective, and the second and third rows are the corresponding rendered semantic images, where the second row is the image before optimization and the third row is the image after optimization. The yellow box indicates the difference comparison with clear semantic segmentation boundaries in the corresponding area.}
	\label{semantic}
\end{figure*}

\section{Experiment and evaluation}
\subsection{Experimental Setup}
\textit{Datasets}: We comprehensively evaluated the proposed method on both synthetic and real-world datasets, including 8 sequences from the Replica dataset\cite{straub2019replica}, 6 sequences from the ScanNet dataset\cite{ dai2017scannet}.

\textit{Metrics}: Following the evaluation section of NEDS-SLAM\cite{ji2024neds}, we use RSNR (Peak Signal-to-Noise Ratio), SSIM (Structural Similarity)\cite{ wang2004image}, and LPIPS (Learned Perceptual Image Patch Similarity)\cite{zhang2018unreasonable} for evaluating RGB reconstruction quality. For depth reconstruction quality, we use the L1. For semantic reconstruction quality, we use mIoU (mean Intersection over Union). For camera localization accuracy, we use ATE Mean and ATE RMSE.

\textit{Baselines}: We selected several NeRF-based SLAM systems, including NICE-SLAM\cite{zhu2022nice}, Vox-Fusion\cite{yang2022vox}, Co-SLAM\cite{wang2023co}, ESLAM\cite{johari2023eslam}, NIDS-SLAM\cite{haghighi2023neural}, DNS-SLAM\cite{li2023dns}, and SNI-SLAM\cite{zhu2024sni}, for comparison with our method. Additionally, we chose 3D GS-based SLAM systems, such as Splatam\cite{keetha2024splatam}, Photo-SLAM\cite{huang2024photo}, NEDS-SLAM\cite{ji2024neds}, and SGS-SLAM\cite{li2025sgs}, to compare with our approach. All comparative data in this paper are derived from the original texts of the aforementioned baselines.

\textit{Platform}: The hardware platform used for the experiments is a laptop equipped with an NVIDIA RTX 3060 GPU and an AMD Ryzen 7 5800H CPU. The software platform is Ubuntu 18.04, with the code written in C++. For convenience of reimplement, we have created a docker container for the code and dependencies.

\textit{Parameters}: We set the number of image pyramid levels to 3. We set ${\lambda _r}=0.2$ and ${\lambda _s}=0.2$.

\subsection{Quantitative Experiments}
Table.\ref{t1} shows the quantitative comparison of RGB reconstruction quality between our method and the baselines on 8 sequences of the Replica dataset. As can be seen, our proposed method performs well in RGB reconstruction quality, especially in PSNR and LPIPS metrics, achieving the best results and surpassing the current state-of-the-art methods. Compared to the second-best results, our method improves by 11.13\% in PSNR and 68.57\% in LPIPS. This improvement is due to the introduction of 3D multi-level pyramid gaussian splatting in our method, which better restores the scene details compared to SGS-SLAM\cite{li2025sgs} and Photo-SLAM\cite{huang2024photo}. Our method also achieves competitive second-best performance in SSIM.

Table.\ref{t2} shows the average quantitative comparison of Depth, ATE, and FPS metrics between our method and the baselines on 8 sequences of the Replica dataset. Our method demonstrates competitive performance in both depth and FPS metrics. The performance of ATE is close to Photo-SLAM\cite{huang2024photo}, as we directly use the tracking module of ORB-SLAM3\cite{campos2021orb} without further optimization. Our method also achieves better performance of Tracking FPS and Mapping FPS compared with SGS-SLAM\cite{li2025sgs}(implement with Python code), which enables our system to run in real-time. 

Table.\ref{t3} shows the quantitative comparison of semantic image reconstruction quality between our method and the baselines on 4 sequences of the Replica dataset. Compared to the currently best-performing SGS-SLAM\cite{li2025sgs}, our method achieves a higher average mIoU of 94.32. 

\subsection{Qualitative Experiments}
Fig.\ref{RGB} shows the qualitative results of randomly rendered RGB images on 8 sequences of the Replica dataset. It can be seen that our method accurately restores fine details in the scene, such as small numbers, textures, and boundaries.

Additionally, Fig.\ref{depth} shows the qualitative comparison results between rendered depth images and groundtruth depth images for our method on the office0 sequence of Replica dataset. It is worth mentioning that even though the input depth image has missing areas, our method is still able to render the depth information in these regions, maintaining good consistency with the surrounding depth information.

Furthermore, Fig.\ref{semantic} shows the qualitative comparison results of semantic image rendering on 4 sequences of the Replica dataset. Our method significantly restores the semantic segmentation results of the scene, especially at the boundaries. The comparison before and after optimization further demonstrates the effectiveness of our proposed semantic image rendering and optimization method.

\begin{table}[]
\caption{Quantitative comparison of semantic image reconstruction quality between our method and baselines on 4 sequences of  Replica dataset.}\label{t3}
\resizebox{\linewidth}{!}{
\begin{tabular}{cccccc}
\hline
Method     & AVG.mIoU(\%)↑   & room0          & room1          & room2          & office0        \\ \hline 
NIDS-SLAM\cite{haghighi2023neural}  & 82.37          & 82.45          & 84.08          & 76.99          & 85.94          \\ [5pt]
DNS-SLAM\cite{li2023dns}   & 84.77          & 88.32          & 84.90           & 81.20           & 84.66          \\[5pt]
SNI-SLAM\cite{zhu2024sni}   & 87.41          & 88.42          & 87.43          & 86.16          & 87.63          \\[5pt]
NEDS-SLAM\cite{ji2024neds}  & 90.78          & 90.73          & 91.20           & /              & 90.42          \\[5pt]
SGS-SLAM\cite{li2025sgs}   & {\ul 92.72}          & \textbf{92.95} & {\ul 92.91}          & {\ul 92.10}           & {\ul 92.90}           \\[5pt]
RGBDS-SLAM(Ours) & \textbf{94.32} & {\ul 92.67}          & \textbf{95.77} & \textbf{94.91} & \textbf{93.91} \\ 
\hline
\end{tabular}%
}
\end{table}

\subsection{Ablation Study}
\begin{figure*}[ht]
	\centering
	\includegraphics[scale=0.5]{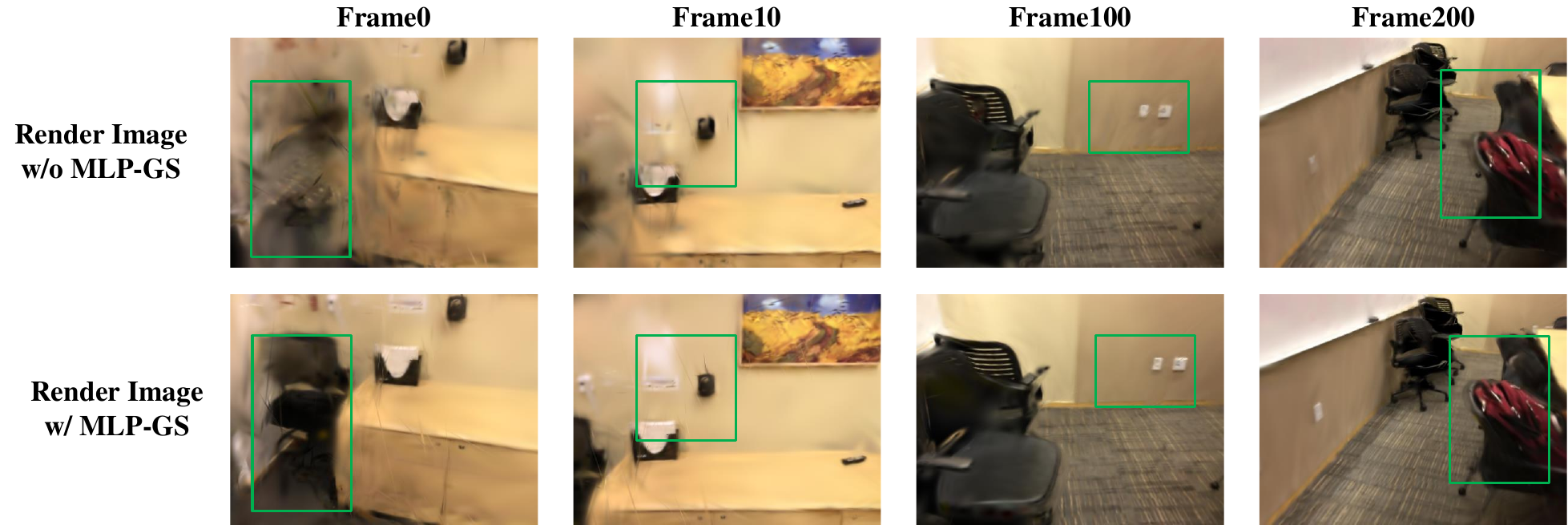}
	\caption{Ablation study of the multi-level pyramid gaussian splatting in our proposed method on ScanNet dataset. The first row shows the multi-frame RGB image rendering results using the standard GS process instead of our proposed MLP-GS. The second row shows the corresponding multi-frame RGB image rendering results using MLP-GS. The areas with significant differences in the images are highlighted with green boxes.}
	\label{ablation}
\end{figure*}

\textit{Effectiveness of MLP-GS Module}:
Fig.\ref{ablation} shows the ablation study of the multi-level pyramid gaussian splatting module in our proposed method on ScanNet dataset. It can be seen that the rendered images using the MLP-GS process clearly preserve more scene details, including object contours, boundaries between objects, and the fine-grained details of small objects.

\begin{table}[]
\caption{Ablation study of the tightly-coupled multi-feature reconstruction optimization mechanism in our proposed method.}\label{t4}
\resizebox{\linewidth}{!}{
\begin{tabular}{cccccc}
\hline
Method                & PSNR↑          & SSIM↑          & LPIPS↓         & Depth↓         & mIoU↑          \\ \hline
w/o depth \& semantic & 36.62          & 0.950          & 0.050          & /              & /              \\ [5pt]
w/o depth             & 38.36          & {\ul 0.966}    & {\ul 0.035}    & /              & {\ul 94.20}    \\ [5pt]
w/o semantic          & {\ul 38.44}    & 0.965          & 0.040          & {\ul 0.345}    & /              \\ [5pt]
w/ depth \& semantic  & \textbf{38.85} & \textbf{0.967} & \textbf{0.035} & \textbf{0.342} & \textbf{94.32} \\ \hline
\end{tabular}%
}
\footnotesize{  \\ w/o means without, w/ means with.}
\end{table}

\textit{Effectiveness of TCMF-RO Module}: Table.\ref{t4} shows the ablation study of the tightly-coupled multi-feature reconstruction optimization module in our method, which focuses on the impact of depth and semantic features on various metrics. As can be seen, when both depth and semantic features are included in the optimization, the best performance is achieved. This demonstrates the effectiveness of our proposed tightly-coupled multi-feature reconstruction optimization mechanism, where RGB, depth, and semantic features mutually promote each other, leading to an overall improvement in the reconstruction quality.

\begin{figure}[ht]
	\centering
	\includegraphics[scale=0.47]{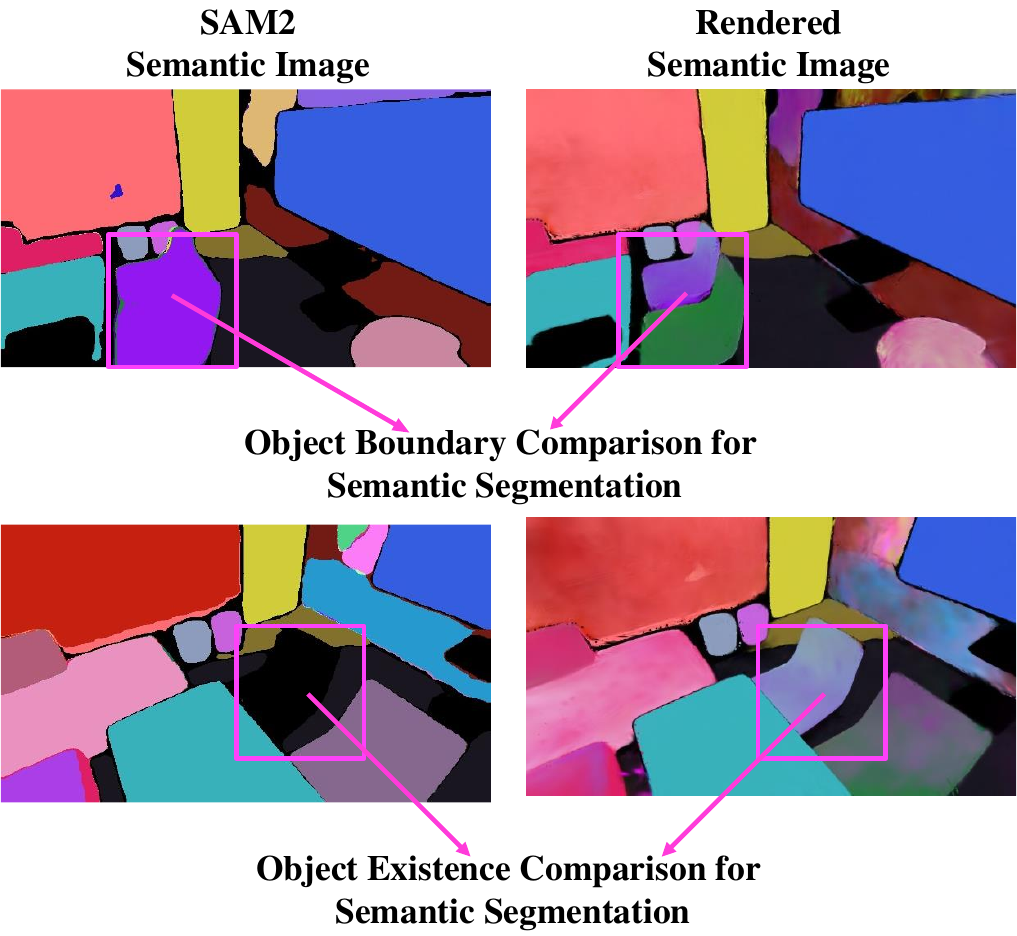}
	\vspace{-2mm}
	\caption{Comparison between the SAM2 segmentation results and the rendered results after our method performs semantic reconstruction. The first row displays a comparison of object boundaries in the semantic segmentation, while the second row shows a comparison of object existence in the semantic segmentation.}
	\label{ablation2}
	\vspace{-7mm}
\end{figure}

\textit{Correction of Semantic Information}: In the above experiments, we only used the groundtruth semantic images for training from the Replica dataset. The current best-performing SGS-SLAM \cite{li2025sgs} also relies solely on groundtruth semantic images for evaluation. However, since groundtruth semantic images are difficult to obtain and cannot be scaled to real-world scenarios, we used the SAM2 network\cite{ravi2024sam} to obtain semantic segmentation results and replaced the original groundtruth semantic images for our experiments. Fig.\ref{ablation2} shows a comparison between the SAM2 segmentation results and the rendered results after semantic reconstruction results of our method. We observed that, compared to semantic groundtruth, the SAM2 segmentation results lack consistency and continuity, with many instances of missed and incorrect segmentation. However, our method does not directly optimize based on the SAM2 segmentation results; instead, it uses multi-frame observations to correct the semantic information, which addresses issues like unclear object boundaries and object omissions in the segmentation. It demonstrates that our proposed method is scalable and can be easily extended to real-world applications.

\section{Conclusion}
In this paper, we propose RGBDS-SLAM, which is a  complete RGB-D semantic dense SLAM system, focusing on gaussian mapping. We first introduce a 3D multi-level pyramid gaussian splatting method to reconstruct the details and consistency of the scene. We futhermore design a tightly coupled multi-feature reconstruction optimization mechanism that promotes the optimization of RGB, depth, and semantic features, enhancing each other. Experiments also demonstrate the effectiveness and scalability of our proposed method. However, we have not considered the issue of dynamic scenes. Robustly reconstructing the RGB, depth, and semantic information in dynamic scenes will be the focus of our future work.

\bibliographystyle{IEEEtran}
\bibliography{ref}



\end{document}